\documentclass[sigconf]{acmart}

\usepackage{booktabs} 

\copyrightyear{2019} 
\acmYear{2019} 
\setcopyright{acmlicensed}
\acmConference{}{}{}
\acmBooktitle{}
\acmPrice{}
\acmDOI{}
\acmISBN{}
\usepackage{multirow,tabularx}

\DeclareMathOperator{\igd}{\mbox{IGD}}
\DeclareMathOperator{\igdx}{\mbox{IGDX}}

\usepackage[ruled]{algorithm2e}
\SetCommentSty{mycommfont}

\newcommand{\inX}{\in \cX}
\newcommand{\hvc}{hill-valley clustering}

\newcommand{\bbC}{\mathbb{C}}
\newcommand{\bbE}{\mathbb{E}}

\newcommand{\bbK}{\mathbb{K}}

\newcommand{\bbR}{\mathbb{R}}

\newcommand{\bc}{\mathbf{c}}

\newcommand{\bff}{\mathbf{f}}

\newcommand{\bx}{\mathbf{x}}

\newcommand{\by}{\mathbf{y}}

\newcommand{\cA}{\mathcal{A}}

\newcommand{\cC}{\mathcal{C}}

\newcommand{\cE}{\mathcal{E}}

\newcommand{\cO}{\mathcal{O}}
\newcommand{\cP}{\mathcal{P}}

\newcommand{\cX}{\mathcal{X}}

\newcommand{\floor}[1]{\left\lfloor #1 \right\rfloor}

\newcommand{\norm}[1]{\left\lVert #1 \right\rVert} 

\makeatletter
\def\Ddots{\mathinner{\mkern1mu\raise\p@
\vbox{\kern7\p@\hbox{.}}\mkern2mu
\raise4\p@\hbox{.}\mkern2mu\raise7\p@\hbox{.}\mkern1mu}}
\makeatother

\newcommand{\clusteralg}{hill-valley clustering}
\newcommand{\alg}{MO-HillVallEA}

\newcommand{\mam}{MAMaLGaM}

\begin{document}
\title[Real-Valued Evolutionary MMMO by {\clusteralg}]{Real-valued Evolutionary Multi-modal Multi-objective Optimization by Hill-valley Clustering}

\author{S.C. Maree}
\affiliation{
\institution{Amsterdam UMC \\ University of Amsterdam}
\country{Amsterdam, The Netherlands}
}
\email{s.c.maree@amc.uva.nl}

\author{T. Alderliesten}
\affiliation{
\institution{Amsterdam UMC \\ University of Amsterdam}
\country{Amsterdam, The Netherlands}
}
\email{t.alderliesten@amc.uva.nl}

\author{P.A.N. Bosman}
\affiliation{\institution{Centrum Wiskunde \& Informatica}
\country{Amsterdam, The Netherlands}}
\email{peter.bosman@cwi.nl}

\renewcommand{\shortauthors}{Maree et. al.}

\begin{abstract}
In model-based evolutionary algorithms (EAs), the underlying search distribution is adapted to the problem at hand, for example based on dependencies between decision variables. Hill-valley clustering is an adaptive niching method in which a set of solutions is clustered such that each cluster corresponds to a single mode in the fitness landscape. This can be used to adapt the search distribution of an EA to the number of modes, exploring each mode separately. Especially in a black-box setting, where the number of modes is a priori unknown, an adaptive approach is essential for good performance. In this work, we introduce multi-objective hill-valley clustering and combine it with MAMaLGaM, a multi-objective EA, into the multi-objective hill-valley EA (MO-HillVallEA). We empirically show that MO-HillVallEA outperforms MAMaLGaM and other well-known multi-objective optimization algorithms on a set of benchmark functions. Furthermore, and perhaps most important, we show that MO-HillVallEA is capable of obtaining and maintaining multiple approximation sets simultaneously over time.
\end{abstract}

%
%




\maketitle

\section{Introduction}
A multi-objective optimization problem comprises two or more objective functions that need to be optimized simultaneously. When these objectives are conflicting, instead of a single optimal solution, multiple Pareto-optimal solutions exist. Without further information, none of these solutions is better than any other. The aim of multi-objective optimization is to obtain as many diverse Pareto-optimal solutions as possible to be presented to the user for decision making. Multi-objective evolutionary algorithms (MOEAs) are aimed to find a set of Pareto-optimal solutions with different trade-offs in the objectives \cite{bosman10,deb02,zhang07}. Distinct solutions that have (almost) equivalent objective values often get lost during optimization, as there is no mechanism to maintain all of these solutions \cite{bosman10,deb02}. Maintaining these could however provide insight that can be used during decision making, and could improve performance.

The aim of multi-modal multi-objective optimization is to obtain a good approximation of the set of \textit{all} Pareto-optimal solutions  \cite{tanabe18,li17,yue17}. In other words, while multi-objective optimization is aimed to approximate the \textit{Pareto front} in objective space, multi-modal multi-objective optimization is aimed to approximate the\textit{ Pareto set} in decision space.

Recently, a number of MOEAs for multi-modal optimization (MMOEAs) have been introduced, by applying niching techniques to existing MOEAs. These MMOEAs are then aimed to approximate the Pareto set with a diverse set of high-quality solutions in decision space. DN-NSGA-II is a niched NSGA-II using crowding, which shows better decision-space diversity compared to standard NSGA-II \cite{liang16}. MO\_Ring\_PSO\_SCD is a niched particle swarm optimizer using an index-based ring topology \cite{yue17}. MOEA/D-AD is a niched MOEA/D, which was shown to outperform MO\_Ring\_PSO \_SCD in terms of decision-space diversity \cite{tanabe18}.

The contribution of this work is twofold. We introduce a novel niching technique, {\hvc}, for multi-objective optimization in Section~\ref{sec:hillvalley}. Hill-valley clustering, previously introduced for single-objective optimization \cite{maree18,maree18b}, is an adaptive clustering method in which a set of solutions is clustered into niches by spending additional function evaluations to determine whether two solutions belong to the same niche. Second, in Section~\ref{sec:mamalgam}, we combine {\hvc} with the MOEA {\mam} \cite{rodrigues14} into an MMOEA, which we refer to as {\alg}. In Section~\ref{sec:experiments}, {\alg} is benchmarked against {\mam} and other (M)MOEAs on different (multi-modal) multi-objective optimization problems. We discuss this work in Section~\ref{sec:discussion}, and conclude in Section~\ref{sec:conclusion}.



\section{Multi-objective niching}
\label{sec:hillvalley}
We define multi-modal multi-objective optimization by a to-be-minimized function $\bff : \cX \rightarrow \bbR^m$, where $\bff = [f_0,\ldots, f_{m-1}]$ is an $m$-dimensional objective function and where $\cX$ is the $n$-dimensional decision space $\cX\subseteq \bbR^n$.
A solution $\bx\inX$ is said to \textit{dominate} another solution $\by\inX$, when $\bx$ is better than $\by$ in at least one objective, and not worse in the others. The set of all non-dominated solutions is called the \textit{Pareto set}. The image of the Pareto set under $\bff$ is called the \textit{Pareto front}. The aim of multi-objective optimization is to find an approximation set $\cA$ of solutions that approximates the Pareto front, while the aim of multi-modal multi-objective optimization is to approximate the Pareto set. 

In optimization, a \textit{niche} is a subset of the decision space where only one mode resides. In this work, we consider multi-objective niching as the partitioning of the decision space into the minimum number of niches required so that all objectives are uni-modal within a niche. Each niche then contains one local Pareto set. The Pareto set is a subset of the union of all local Pareto sets. When a local Pareto set maps to (part of) the global Pareto front, we refer to it as a global Pareto set. For a formal definition of local Pareto sets, or locally efficient sets, we refer the reader to \cite{Kerschke17}.

\begin{figure}\hspace*{-0.8cm}    \vspace*{-0.5cm}           
\includegraphics[width=1.1\columnwidth]{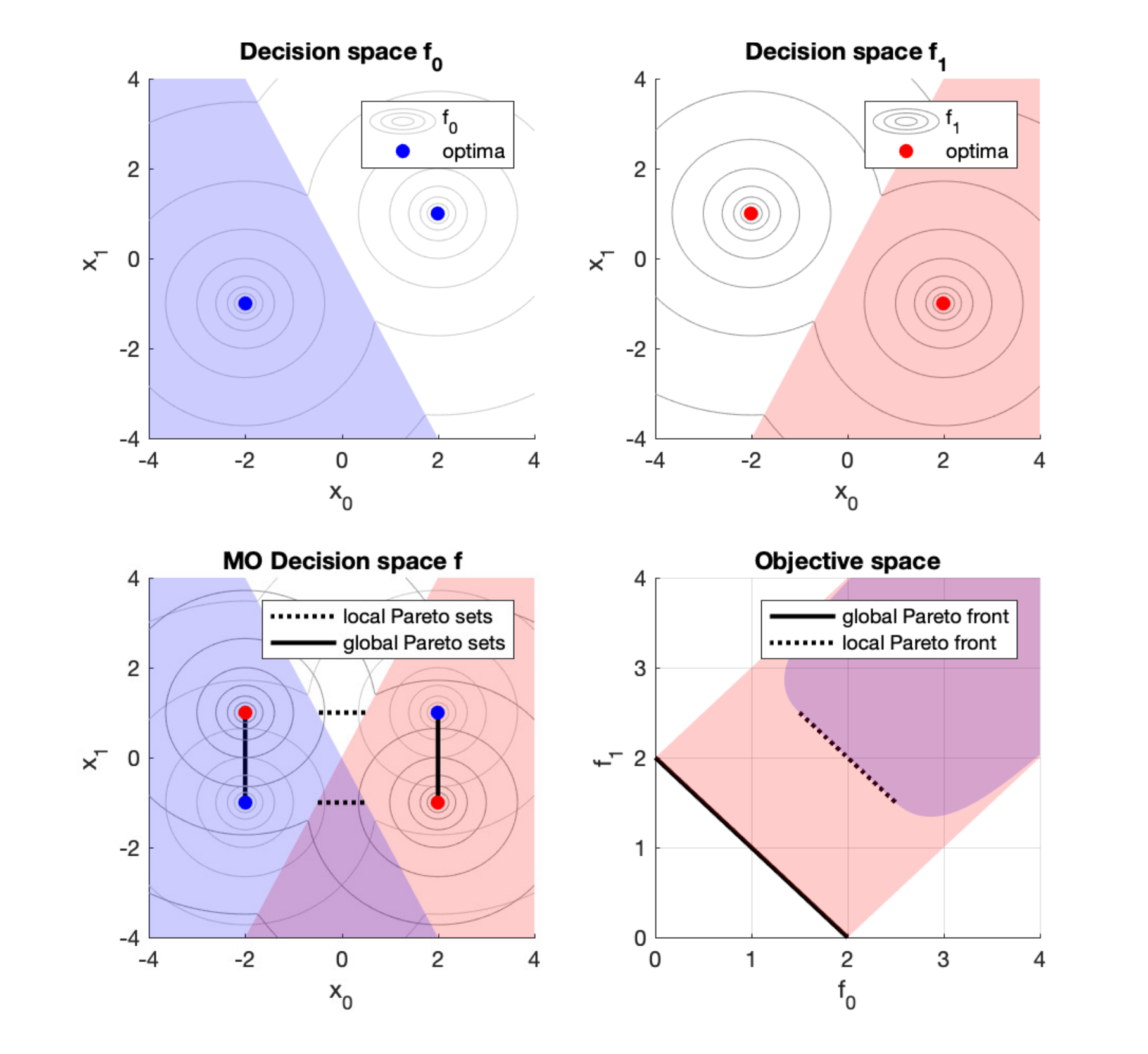}
\caption{Niches for the MinDist problem with two objectives. The top subfigures show the single-objective niches for both objectives in different colors. The lower left subfigure show the four niches for the multi-objective problem $\bff = [f_0,f_1]$, together with the four local Pareto sets. The lower right subfigure shows the objective space, where the red and purple niches are shown. The blue niche overlaps exactly with red niche, and white with purple.}
\label{fig:mo_niches}
\end{figure}

Let us demonstrate this using the minimum distance (MinDist) problem \cite{bosman10, ishibuchi10}, which is based on multiple objective functions of the form, $f_i(\bx|\bc_0, \bc_1)  = \min\{\norm{\bx-\bc_0}, \norm{\bx-\bc_1} \}, $ for center points $\bc_0,\bc_1 \inX$. For the MinDist problem with two objectives, we use the objectives $\bff = [f_0(\bc_0,\bc_1), f_1(\bc_2,\bc_3)]$ and center points  $\bc_0 = [ -2,-1,0,\ldots, 0]$, $\bc_1 = [ 2,1,0,\ldots, 0]$, $\bc_2 = [ -2,1,0,\ldots, 0]$, and $\bc_3 = [ 2,-1,0,\ldots, 0]$ and the Euclidean distance. MinDist with $m = 2$ objectives is visualized in Figure~\ref{fig:mo_niches}. The overlapping niches in the two objectives result in four multi-objective niches. Two local Pareto sets map to the global Pareto front, while the other two map to local Pareto fronts. A local Pareto set can be understood as a line in the decision space, connecting optima of different objectives. However, parts of these the lines resulting in the local Pareto sets are dominated by the global Pareto sets. Note that, even though both objectives have only global optima, the multi-objective problem has both local and global Pareto sets.

\subsection{Single-objective {\hvc}}
Hill-valley clustering is a niching approach that was introduced for single-objective multi-modal optimization \cite{maree18,maree18b}. It can be used to cluster a set of solutions such that each cluster resides in a single niche. To determine whether two solutions belong to the same niche, the \textit{hill-valley test} is used \cite{ursem99,maree18}. The idea behind this test is that, when there is a \textit{hill} between two solutions, they belong to different \textit{valleys} (niches). For this, a number of solutions are evaluated along the line segment connecting the two solutions in decision space. If any of these test solutions has worse fitness than the two solutions, the two solutions belong to different niches.

To cluster an entire population of solutions, an approach inpsired by nearest-better clustering \cite{preuss10} is used. First, the population is sorted on fitness, starting with the best solution in the population, which forms the first cluster. Then, iteratively, each next-best solution is tested to determine whether it belongs to the niche of the nearest solution that has better fitness. When a solution-pair is rejected, the next-nearest solution with better fitness is considered. It could be that the nearest solution belongs to a different niche, but a next-nearest solution that is located slightly further, but in a different direction, belongs to the same niche. If a solution does not belong to the same niche as its $N_n = n + 1$ nearest-better neighbors, a new cluster is formed from that solution. This is repeated until the entire population is clustered. A detailed description of single-objective {\hvc} can be found in \cite{maree18}.

\begin{algorithm}
\SetKwData{Sort}{Sort}
\SetKwData{Find}{Find}
\SetKwData{Add}{Add}
\SetKwInOut{Function}{function}
\SetKwInOut{Input}{input}
\SetKwInOut{Output}{output}
\Function{$[$boolean$] = $ HillValleyTest($\bx, \by, N_t, f$)}
\Input{Solutions $\bx, \by$, Number of test points $N_t$, Objective function $f$ \small{(to be minimized)}}
\Output{$\bx, \by$ belong to the same niche? \small{(boolean)}}
\BlankLine
\For{$k = 0,\ldots,N_t-1$ }
{
$\bx_k = \bx + \frac{k}{N_t + 1}(\bx-\by)$\;
\If{$\max\{f(\bx),f(\by)\} < f(\bx_k)$}{
return false\;
}
}
return true\;
\hrulefill

\Function{$\bbC$ = HillValleyClustering($\cP, \mathbf{f}$)}
\Input{Population of solutions $\cP$,\\ Objective function $\mathbf{f} : \bbR^n \rightarrow \bbR^m$}
\Output{Set of clusters $\bbC = \{\cC_0,\ldots,\cC_{s-1}\}$ }
\BlankLine
$\Delta = \sqrt[n]{V_\cP/|\cP|}$\tcp*{$V_\cP$ = volume of bounding box of $\cP\subset\bbR^n$}
\For{$l = 0,\ldots,m-1$}
{
\BlankLine
\tcp{Single-objective {\hvc} for objective $l$}
\Sort $\bx\in\cP$ on fitness value of $l$-th objective, fittest first\;
$\cC = \{\bx_0\}$; $\bbK_l = \{\cC\}$\tcp*{Init cluster from best solution}
\BlankLine

\For{$i = 1,\ldots,|\cP|-1$}
{
	\tcp{Compute Euclidean distances to all better $\bx_p$}
	$\delta_p = \norm{\bx_i - \bx_p}, \mbox{ for } p = 0,\ldots,i-1$\;
	
	\BlankLine
	\For{$j = 0,\ldots,\min\{i-1,n\}$}
	{
	\BlankLine
		$k$ = index of $j$-th nearest better solution according to $\{\delta_p\}_{p=0}^{i-1}$;
		
		\BlankLine

		\If{HillValleyTest$\left(\bx_{k}, \bx_i, 1 + \floor{ \delta_k / \Delta}, f_l\right)$}
		{
			$\cC_{(\bx_k)} = \cC_{(\bx_k)} \cup \{\bx_i\}$\tcp*{Add to cluster of $\bx_{k}$}
			break\;
		}
	}
	
	\If{$\bx_i$ was not added to any cluster}
	{
	$\cC := \{\bx_i\}$; $\bbK_l = \bbK_l \cup \cC$ \tcp*{Init new cluster}
	}
}

}

$\bbC = \bigcap\{\bbK_l\}_{l=0}^{m-1}$\tcp*{all intersections of all clusters in $\bbK_l$}

\caption{Hill-valley Clustering}
\label{alg:hill_valley}
\end{algorithm}


\subsection{Multi-objective {\hvc}}
We extend {\hvc} to the multi-objective case using insight gained from Figure~\ref{fig:mo_niches}. Single-objective {\hvc} is performed once for each objective. The result is $m$ (different) clusterings of the population. These are then reduced to a single cluster set by taking intersections, similar as the differently colored regions illustrated in Figure~\ref{fig:mo_niches}. That is, solutions that belong to the same niches for all objectives are grouped together. See pseudo code in Algorithm~\ref{alg:hill_valley}.

To improve efficiency, test solutions evaluated by the hill-valley test are stored to prevent evaluating them twice for the clustering of different objectives. Furthermore, test solutions for solution-pairs that ended up in the same cluster are added to that cluster as well. Test solutions for solution-pairs that ended up in different clusters are discarded, as it is not clear to which cluster they should be added without further testing.

Hill-valley clustering uses at least $N$ function evaluations to cluster a population of size $N$, as at least one neighbor needs to be checked for each solution. An upper bound is more difficult to formulate because the number solutions used in the hill-valley test depends on the distance between nearest better solutions, which is problem- and sample-dependent. As at most $n+1$ neighbors are considered per solution, the number of function evaluations is roughly $\cO((n+1)N)$ in practice. 

\section{{\alg}}
\label{sec:mamalgam}

We use {\hvc} to construct a multi-modal multi-objective evolutionary algorithm that we refer to as the multi-objective hill-valley evolutionary algorithm ({\alg}). The idea behind the algorithm is that every generation, the population is clustered using {\hvc}. Then, each cluster is explored with a generation of a core optimization algorithm. For this we use different versions of {\mam} in this work. {\mam} is an estimation-of-distribution algorithm, i.e., a type of model-based evolutionary algorithm that learns a probability distribution to subsequently sample new solutions from \cite{rodrigues14}. 

The population $\cP$ is initialized by uniform sampling, after which {\hvc} is used to obtain  clusters $\bbC = \{\cC_0,\cC_1,\ldots\}$. For each cluster $\cC_i$, a generation of the core optimization algorithm is performed to generate offspring $\cO_i$. Core optimization algorithms of the model-building type, such as {\mam}, generally estimate model-parameters over the course of multiple generations. Therefore, a set of model parameters $\rho_i$ is maintained for each cluster $\cC_i$. To smoothly transfer model parameters over generations, clusters are linked to the nearest cluster from the previous generation. This distance is measured by the Euclidean distance between the cluster means in decision space. Note that this supports having a variable number of clusters every generation. 

Besides the main population, an elitist archive $\bbE$ is maintained that contains the best solutions over time. To allow for multi-modal optimization, all non-dominated solutions within a niche need to be maintained, even if they are dominated by a solution from a different niche. Therefore, a local elitist archive, or subarchive, $\cE_i$ is generated from each cluster $\cC_i$ by maintaining only the non-dominated solutions within that cluster. The (global) elitist archive $\bbE$ is then the set of all subarchives, i.e., $\bbE = \{\cE_0, \cE_1, \ldots \}$. Note that in this way, dominated solutions can end up in the elitist archive, but only if the dominating solution is from a different subarchive, and thus presumably from a different niche. In order to incorporate subarchives from the previous generation into the clusters of the next generation, one must know to which cluster each subarchive corresponds. Therefore, all elites are added to the population before {\hvc} is applied. In that way, clusters contain all new offspring and all elites. After the clustering process and the construction of the elitist archive, the elites are removed from the clusters, but maintained in the elitist archives. Note that {\mam}, within {core\_opt\_generation}($\cC,\cE,\rho$), does again add a few elites back in the clusters to improve convergence.

To reduce the number of function evaluations spent during {\hvc}, whenever two elites are considered in the hill-valley test that originate from the same subarchive, they are said to be part of the same niche, without further testing. However, when two elites originate from different subarchives, the hill-valley test is performed. Due to the discrete nature of the hill-valley test, solutions close to the boundaries of a niche can be clustered incorrectly, resulting in small, low-fitness clusters. By testing elites from different niches every generation, these clusters are more often merged with the correct neighboring clusters.

If, at the end of the generation, the elitist archive size exceeds a user-defined target size $N_\bbE$, adaptive objective-space discretization is performed, to reduce computational cost while maintaining diversity in the archive \cite{rodrigues14, luong12}. All subarchives are discretized using the same grid size, which is adapted until the total archive size is less than the given target size $N_\bbE$. Note that this maintains diversity within subarchives, but does not focus on total diversity over all archives. The main generational loop for {\alg} is inspired by {\mam}, and pseudo code is given in Algorithm~\ref{alg:hillvallea}.

\begin{algorithm}
\SetKwInOut{Function}{function}
\SetKwInOut{Input}{input}
\SetKwInOut{Output}{output}
\SetKwData{Sort}{Sort}
\SetKwData{Find}{Find}
\SetKwData{Add}{Add}
\Function{[$\bbE$] = {\alg}($\mathbf{f}, N, N_\bbE$)}
\Input{Objective function $\mathbf{f}$, Population size $N$, Elitist archive target size $N_\bbE$}
\Output{Elitist archive $\bbE = \{\cE_0,\cE_1,\ldots\}$}

\BlankLine
$\cP = \text{UniformSampling}(N)$\tcp*{Also evaluates solutions}
$\bbC = \text{HillValleyClustering}(\cP, \mathbf{f})$\tcp*{See Algorithm~\ref{alg:hill_valley}}
$\bbE = \text{ConstructLocalElitistArchives}(\bbC, N_\bbE)$\;
$\rho = \text{InitModelParameters}(\bbC)$\tcp*{For each $\cC_i$, modelparams $\rho_i$}
\BlankLine
\While{budget remaining}
{
\BlankLine

$\cP = \bigcup_i \{\cE_i\}$\tcp*{Copy all elites to the population}
\For{$(\cC_i,\cE_i,\rho_i) \in (\bbC,\bbE,\rho)$}
{
	$(\cO_i,\rho_i) = \text{core\_opt\_generation}\left(\cC_i,\cE_i, \rho_i \right)$\;
	$\cP = \cP\, \cup\,\cO_i$\tcp*{Collect offspring $\cO_i$}
}

$\bbC_\text{prev} = \bbC$\tcp*{Backup old clusters}
$\bbC = \text{HillValleyClustering}(\cP, \mathbf{f})$\tcp*{See Algorithm~\ref{alg:hill_valley}}
$\bbE = \text{ConstructLocalElitistArchives}(\bbC, N_\bbE)$\;

$\bbC = \text{RemoveElitesFrom}(\bbC)$

$(\bbC, \rho) = \text{LinkClusters}(\bbC,\bbC_\text{prev},\rho)$\;

}

\caption{\alg}
\label{alg:hillvallea}
\end{algorithm}

\subsection{\mam}
We briefly describe a generation of {\mam}, and necessary adaptations required for its use as a core optimization algorithm.

{\mam} is initialized with a cluster $\cC_i\in\bbC$ as its population, from which offspring are generated. For this, $\cC_i$ is clustered once more. To prevent confusion with {\hvc}, we refer to these subclusters within {\mam} as \textit{subsets}. {\mam} combines two clustering approaches  to obtain a predefined number of subsets $k$. First, for each of the $m$ objectives, a subset is formed by the $N_c$ best solutions in that objective. Second, the population is sorted based on domination rank, and balanced $k$-leader-means (BKLM) clustering \cite{rodrigues14} is used to cluster the $\tau N$ best-ranked solutions into the remaining $k-m$ subsets, where $\tau$ is the selection fraction. This results in a set of $k$ overlapping subsets. A Gaussian distribution is then estimated for each subset, from which new offspring are sampled. For this, mechanisms from different evolutionary algorithms have been used previously, such as, CMA-ES, iAMaLGaM, and AMaLGaM \cite{rodrigues14}. Subsets are registered to subsets from the previous generation and a one-to-one subset registration is applied. Besides sampling new solutions every generation, a maximum of $\tau N$ elites is put back into the population, selected from the elitist archive based on objective-space diversity. 

To initialize {\mam}, the offspring size needs to be set, for which we use $N_i = \floor{N/|\bbC|}$, i.e., we sample an equal number of offspring for each cluster, with the purpose of better maintaining smaller niches. Furthermore, we fix the total number of subsets $k$ over all clusters within {\alg}. For an instance of {\mam} initialized with $\cC_i\in\bbC$ as population, we set the number of subsets $k_i$ by $k_i = \lfloor k\cdot |\cC_i| /\sum_j|\cC_j|\rfloor $, with $k_i \geq 1$. Only when $k_i > m$, $m$ out of $k_i$ subsets are formed by performing single-objective selection. Because $k_i$ can now vary per generation, the one-to-one subset registration that was previously used is no longer applicable. It is therefore replaced by simply registering each subset to the nearest subset from the linked cluster of the previous generation. As subsets are formed based on objective-space distances, subset distances are computed by the distance of the subset means in objective space as well. Additionally, single-objective subsets are linked with each other if they existed in both generations. Note that this approach is significantly faster than the original one-to-one subset registration.

\subsection{Post-processing the approximation set}
In practice, a user can often process only a limited number of solutions during decision making, and the size of the resulting approximation set $\cA$ that an algorithm obtains must thus be limited. Directly restricting the size of the elitist archive $\bbE$ during optimization has a risk of deteriorating performance, especially when the desired number of solutions is small. Therefore, a post-processing step is applied when the desired approximation set size $N_\cA$ is smaller than the elitist archive target size $N_\bbE$. In that case, the approximation set is formed by combining all local elitist archives that contain at least one solution that is non-dominated within the global elitist archive. If the approximation set still exceeds $N_\cA$, a greedy scattered subset selection is performed to reduce the archive size while preserving decision-space diversity as good as possible. This is the same subset selection algorithm as was used for BKLM clustering \cite{rodrigues14}. A similar post-processing step was used for MOEA/D-AD in \cite{tanabe18}, with the difference that {\alg} can maintain dominated solutions as long as they are from different subarchives.

\subsection{Multi-start scheme}
To overcome the need for parameter tuning of the population size $N$, a multi-start scheme is applied, similar to \cite{bouter17b}, where multiple instances of {\alg} are run simultaneously, each with a larger population size. Recursively, after 8 generations of an instance of {\alg} with population size $N$, one generation of an instance with population size $2N$ is performed. The first instance of {\alg} is initialized with population size $N_\text{base} = 10\cdot(1+m)\cdot (1 + \log n)$ and $k_\text{base} = 1 + m$ subsets, where $m$ is the number of objectives and $n$ the decision-space dimensionality. Each subsequent instance has a population size increased by a factor 2 and number of subsets increased by a factor 1.5. The number of subsets increases slower than the population size, so that the subset size increases over time. A single global elitist archive $\bbE$ is maintained for all instances. Whenever a new instance is initialized, it is counted how many solutions the smaller instances contributed to the elitist archive in their most recent generation. An instance that is responsible for less than 25\% of the total contribution is terminated, except for the latest instance.

\section{Experiments}
\label{sec:experiments}
We empirically benchmark {\alg} against comparable (M)MOEAs. For this, {\mam} was implemented in C++, and {\alg} was implemented in the same framework to reduce behavioral differences due to implementation choices. Source code of {\alg} is available at \href{http://github.com/scmaree/MOHillVallEA}{github.com/scmaree/MOHillVallEA}.
Results of other (M)MOEAs were provided by the authors of \cite{tanabe18}. 

\subsection{Test problems}
We consider the MinDist problem with $m = 2$ objectives described in Section~\ref{sec:hillvalley}  with $n = \{10, 20\}$ decision variables, initialized in the box $[-4,4]^n$. Additionally, we consider the MinDist problem with $m = 3$ objectives, given by $\bff = [f_0(\bc_0,\bc_1),f_1(\bc_2,\bc_3),f_2(\bc_4,\bc_5]$ and center points  $\bc_0 = [ -4,-4,0]$ and $\bc_1 = [ 2,2,0 ]$ for the first objective, $\bc_2 = [ -2,-4,0]$ and $\bc_3 = [4,2,0]$ for the second objective, and  $\bc_4 = [ -3,-2,1]$, and $\bc_5 = [ 3,4,1]$ for the third objective, resulting in two triangle-shaped global Pareto sets. Similarly for the $m = 2$ MinDist problem, zeros are appended to the centers if the number of decision parameters is larger.

Furthermore, we consider six bi-objective multi-modal test problems with $n = 2$ decision variables that are frequently considered in multi-modal literature: Two-On-One \cite{preuss06}, SYM-PART\{1,2,3\} \cite{rudolph07}, and SSUF\{1,3\} \cite{yue17}. The Two-On-One problem and the SSUF1 problem both have two symmetrical global Pareto sets that are connected in decision space and that map onto the same Pareto front in objective space. The SYM-PART problems have nine global Pareto sets, where SYM-PART2 has rotated global Pareto sets and SYM-PART3 has non-linear global Pareto sets. The SSUF3 problem has two global Pareto sets that are shifted, and many local Pareto sets. 

All problems have a bounded decision space, and boundary repair is performed when solutions violate the boundary conditions. Initial populations are sampled uniformly in the entire domain.

\subsection{Performance Measures}
We use the inverted generational distance (IGD) \cite{bosman02} to measure objective-space diversity and the IGDX \cite{zhou09} to measure decision-space diversity, which are given by,
\begin{equation}
\label{eqn:igd}
\begin{split}
\igd_{\cA^*}(\cA) &= \frac{1}{|\cA^*|} \sum_{\by\in\cA^*}\min_{\bx\in\cA}\norm{\bff(\bx)-\bff(\by)}, \\
\igdx_{\cA^*}(\cA)& = \frac{1}{|\cA^*|} \sum_{\by\in\cA^*}\min_{\bx\in\cA}\norm{\bx-\by},
\end{split}
\end{equation}
where $\norm{\cdot}$ is the Euclidean norm, and $\cA^*$ is the reference Pareto set, for which 5000 Pareto-optimal solutions are sampled using the analytical expression of the Pareto set. For the Two-On-One problem, an approximation of the Pareto set is used \cite{preuss06}. Both measures should be minimized, and a perfect score of 0 is achieved when each solution in $\cA^*$ is also in $\cA$. Note that for multi-modal fitness landscapes, a low IGD does not imply a low IGDX. E.g., for the MinDist function in Figure~\ref{fig:mo_niches}, $\igd = 0$ can be achieved for $\igdx = 2$, when one of the global Pareto sets is perfectly approximated, while no solution has been obtained in the other set. Best achievable scores for the IGD and IGDX depend on the maximum number of solutions $N_\cA$ desired in the approximation set. To compute achievable limits for these scores, the reference Pareto set is compared to a subset of it that contains $N_\cA$ solutions. This subset is generated using greedy scattered subset selection \cite{bosman10}. Subset selection has been performed with objective-space distances for the IGD and decision-space distances for the IGDX. 

Additionally, we introduce a novel performance measure that we refer to as the \textit{mode ratio} (MR), which is the ratio of attained modes. We say that a Pareto set, or mode, is attained if the IGDX for that mode is smaller than a predefined threshold $\varepsilon$, which we set to $\varepsilon = 0.05$ for problems with two objectives, and $\varepsilon = 0.1$ for problems with three objectives. To compute the MR, the reference Pareto set $\cA^*$ is partitioned up into different modes by clustering it with {\hvc}. The MR should be maximized, and the best score is 1.0, when all modes are attained.

We aim in this work to improve decision space diversity, measured by the IGDX and MR. To indicate how this affects objective space diversity, the IGD is also included. It should be noted, however, that it is not the primary aim of the multi-modal approaches in this work to optimize it.

\subsection{Visualization of {\hvc}}
\begin{figure*} 
\includegraphics[width=\textwidth]{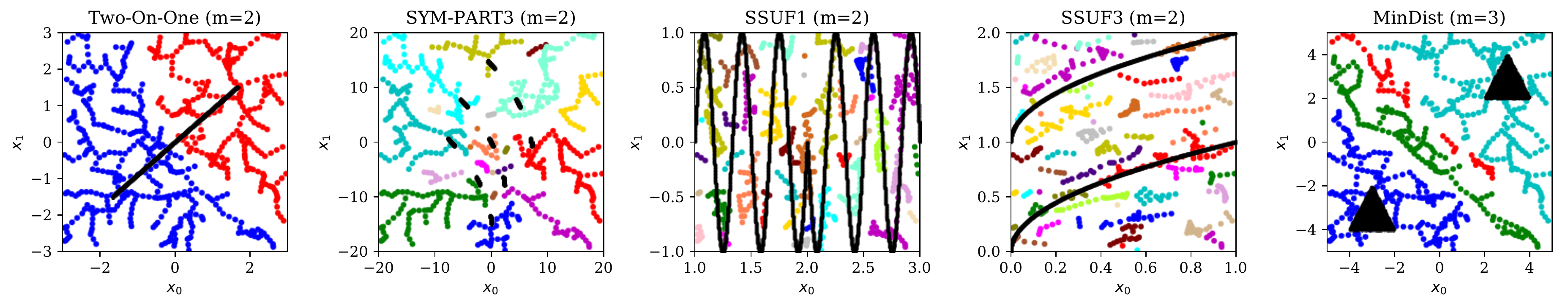}
 \includegraphics[width=\textwidth]{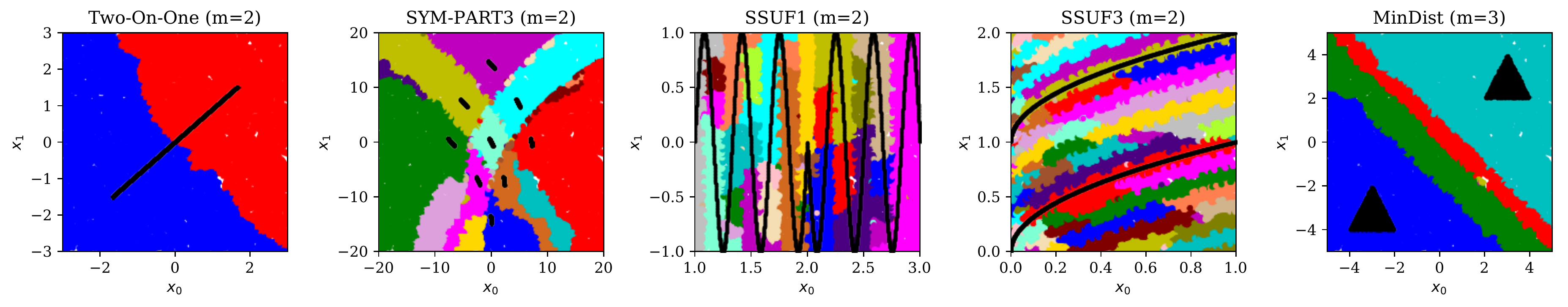}
\caption{Hill-valley clustering of the decision space for different problems, with population size of $N = 250$ (first row) and $N = 10\,000$ (second row). The global Pareto sets are illustrated in black. { For the SYM-PART3 and SSUF problems, the number of clusters obtained in the first row differed from the second row, resulting in a different color mapping.}}
\label{fig:hvc}
\end{figure*}

Hill-valley clustering is visualized for the 2D benchmark problems in Figure~\ref{fig:hvc} with initial population sizes $N = 250$ and $N = 10\,000$. Tree-like clusters are formed because test solutions that are used by the hill-valley test are added to the clusters. A well-structured clustering can be observed for the Two-On-One, SYM-PART3, and three-objective MinDist problems. For the SYM-PART3 problem, low-fitness clusters with only a few solutions are formed on local Pareto sets that exist in between the global Pareto sets, which is caused by the limited number of test points in the hill-valley test. For SSUF1, the domain boundary results in that Pareto sets are connected only by a single point, resulting in more clusters than niches. For the SSUF3 problem, {\hvc} is not always able to connect the long stretched local Pareto sets. This can be reduced by increasing the number of neighbors considers in the clustering process.

\subsection{Comparing core optimization algorithms} 
\label{sec:lops}
We compare different core optimization algorithms and their performance within {\alg} in combination with the multi-start scheme. All parameters of {\alg} and {\mam} are set according to literature \cite{rodrigues14}. For the bi-objective problems, the elitist archive target size is set to $N_\bbE = 1000$ and for the three-objective problems to $N_\bbE = 2500$. To show the full potential of {\alg}, post processing is disabled, by setting the approximation set equal to the union of all subarchives $\cA = \bigcup_i{\cE_i}$.

We compare four versions of {\mam} as core search algorithm. {\mam} (MAM) and {\mam}-univariate (MAMu) estimate a Gaussian distribution with respectively a full-rank and univariate covariance matrix. Similarly, i{\mam} (iMAM) and i{\mam}-univariate (iMAMu) estimate respectively a full-rank and univariate covariance matrix, but the mean and covariance matrix are estimated incrementally over the course of multiple generations. Limiting the search distribution to a univariate covariance matrix or estimating it incrementally typically requires a smaller population size, but results in a worse fit of the fitness landscape.

\subsubsection{Results.}
\begin{figure*}
 \includegraphics[width=0.94\textwidth]{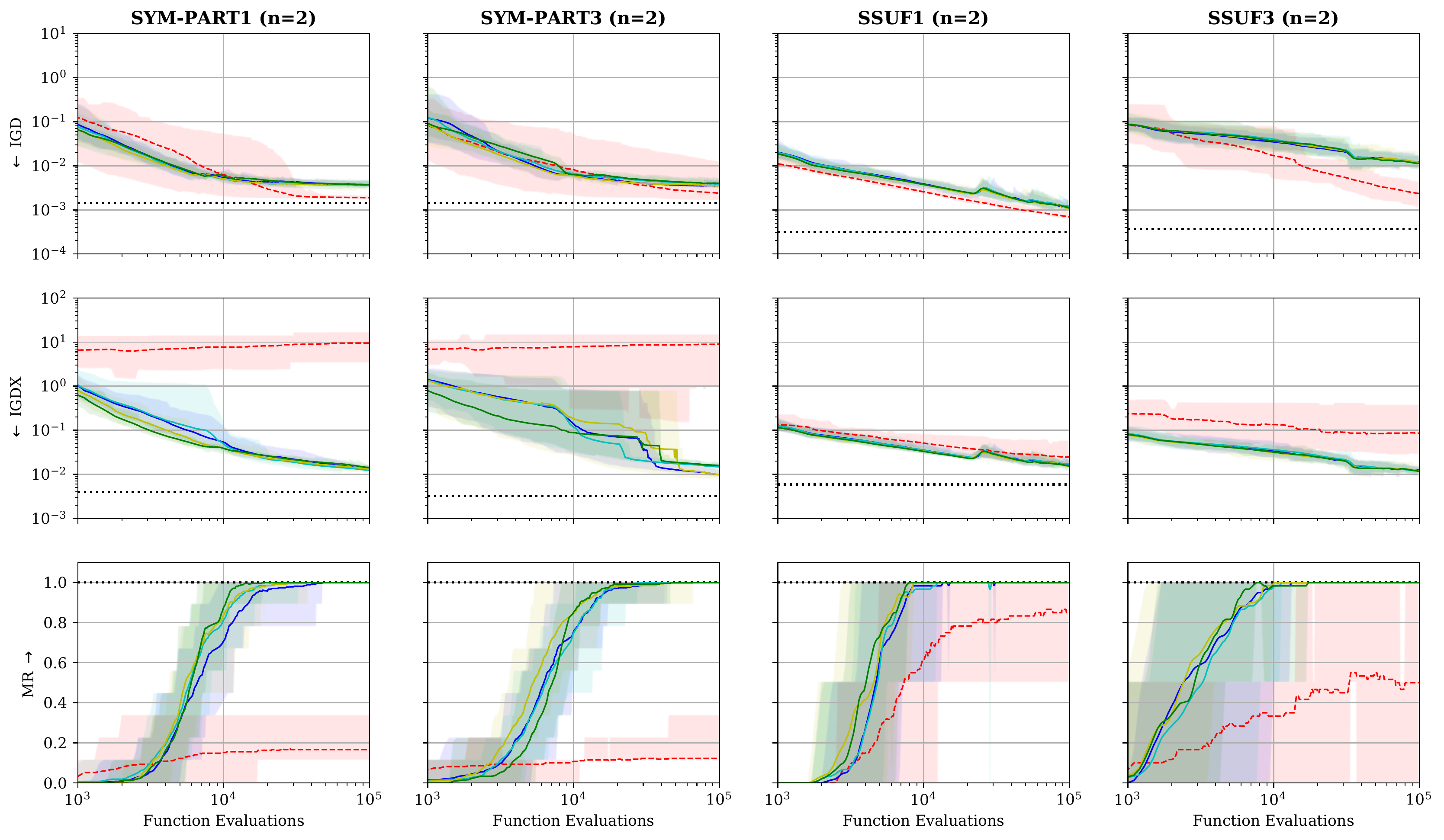} \\ 
 \quad \\
\includegraphics[width=0.94\textwidth]{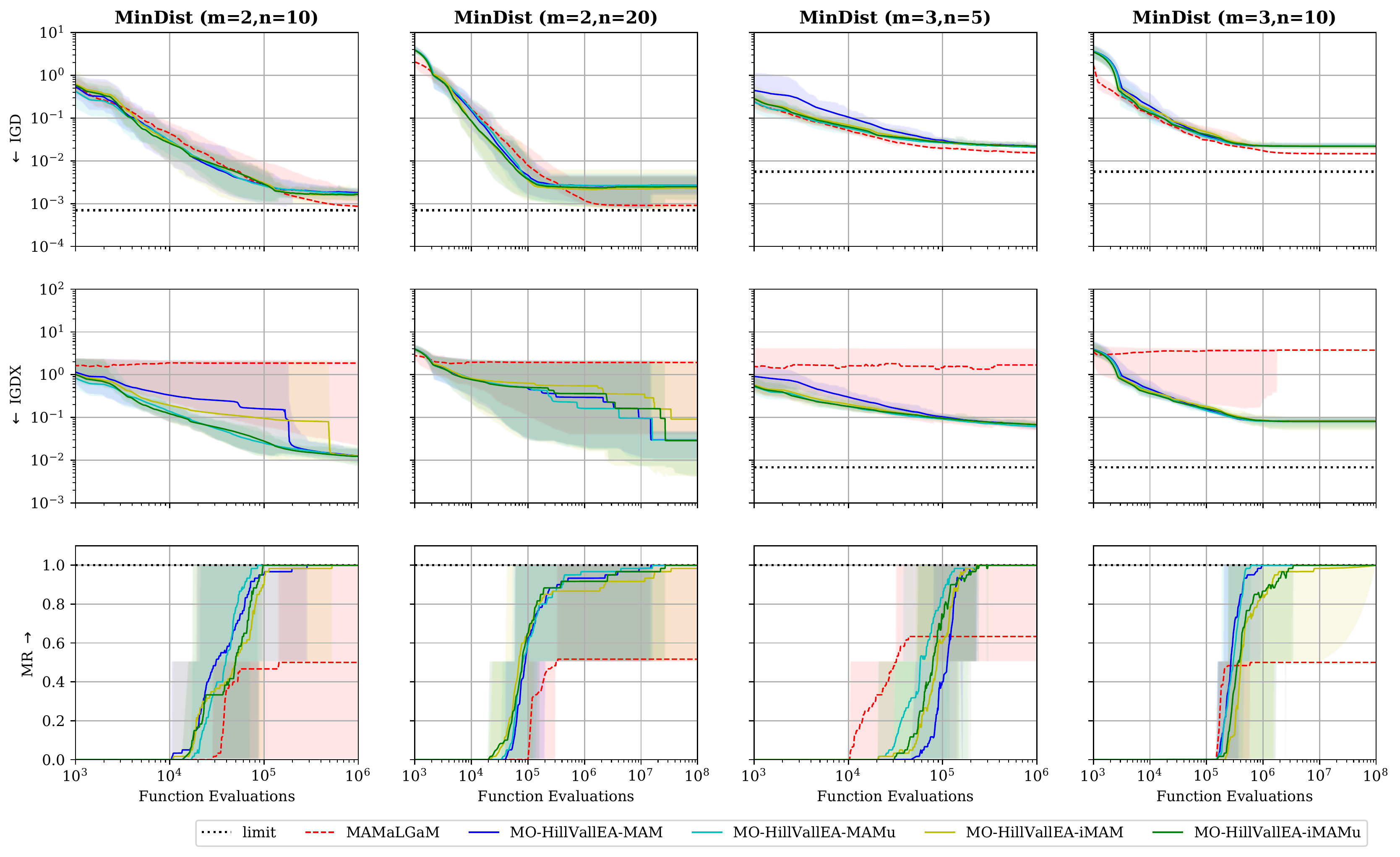}   
\caption{IGD, IGDX, and MR over time for {\alg} with different core optimization algorithms, compared to {\mam}. Results are averaged over 30 runs, shaded areas are min/max scores. Note that the MinDist problems have been run with a larger budget.}
\label{fig:convergence_lops}
 \end{figure*}

Results in terms of the three performance measures for {\alg} with different core optimization algorithms and {\mam} are shown in Figure~\ref{fig:convergence_lops}. In terms of the IGD, all versions of {\alg} perform similar, and a similar rate of convergence can be observed for {\alg} and {\mam}. {\mam} ultimately achieves a better IGD score for all problems, which is mainly due to the limited elitist archive size. When the archive is full, discretization of the archive takes place. {\alg} aims to maintain parameter space diversity within the archive while reducing its size, while {\mam} aims to maintain objective-space diversity. This results in stagnating IGD convergence for {\alg}. Initially, performance of {\alg} on MinDist ($m=2,n=20$) is better in terms of the IGD, which can be attributed to the observation that {\alg} splits the search distribution into two, each focusing on converging locally, while {\mam} tries to approach both modes with a single search distribution, resulting in a slower convergence. As soon as one of the modes is discarded due to generational drift, {\mam} achieves a higher accuracy. In terms of decision-space performance, measured by the IGDX and MR, {\alg} outperforms {\mam} for all problems. This holds especially for SYM-PART and MinDist problems, where {\mam} ultimately obtains only a single mode, while {\alg} obtains all modes. For MinDist ($m=2,n=20$), due to the dimensionality of the decision space, the two modes appear to be close to each other, and distinguishing them becomes hard with a small population size. Therefore, restarts are required to obtain both modes, resulting in the step-like IGDX curves. 

Differences between core optimization algorithms within {\alg} can be observed for the SYM-PART1 and the bi-objective MinDist problems in IGDX and MR. The Pareto sets for these problems are linear and non-rotated, so the univariate MAMu and iMAMu outperform the full-covariance MAM and iMAM, as can be seen in terms of MR performance and a lower IGDX. For SYM-PART3, with rotated Pareto sets, the full-covariance MAM and iMAM outperform the univariate core optimization algorithms by achieving a higher MR. 

 \begin{table*}
 \caption{Mean \textbf{IGD} and \textbf{IGDX} for MMOEAs and MOEAs over 31 runs (standard deviation in brackets). Bold scores are best per problem (not significantly worse than any other). In the \textit{limit} column, best scores achievable with $N_\cA = 100$ are shown.}
\label{tab:igd}
 \begin{center}
 \begin{tabular}{c  ccccc | ccc}
 & & & {\alg} & & MO\_Ring & & &  \\
& Problem &\textit{limit} & -MAM & MOEA/D-AD & \_PSO\_SCD & \multirow{1}{*}{{\mam}} &\multirow{1}{*}{NSGA-II} & \multirow{1}{*}{MOEA/D}\\
 \hline
 \hline
\multirow{6}{*}{\, \rotatebox[origin=c]{90}{\textbf{IGD}} \rotatebox[origin=c]{90}{(objective space)} \,} 
 & Two-On-One & \textit{0.004} & 0.006 \Small{(0.000)} & 0.064 \Small{(0.008)} & 0.061 \Small{(0.006)} & \textbf{0.004} \Small{(0.000)} & 0.049 \Small{(0.002)} & 0.045 \Small{(0.001)}\\
 & SYM-PART1 & \textit{0.018} & 0.047 \Small{(0.009)} & 0.030 \Small{(0.002)} & 0.028 \Small{(0.001)} & \textbf{0.018} \Small{(0.000)} & 0.021 \Small{(0.001)} & 0.047 \Small{(0.001)}\\
 & SYM-PART2 & \textit{0.018} & 0.044 \Small{(0.006)} & 0.031 \Small{(0.002)} & 0.031 \Small{(0.002)} & \textbf{0.018} \Small{(0.000)} & 0.023 \Small{(0.001)} & 0.047 \Small{(0.008)}\\
 & SYM-PART3 & \textit{0.018} & 0.043 \Small{(0.005)} & 0.031 \Small{(0.002)} & 0.032 \Small{(0.003)} & \textbf{0.019} \Small{(0.001)} & 0.023 \Small{(0.001)} & 0.045 \Small{(0.005)}\\
 & SSUF1 & \textit{0.004} & 0.008 \Small{(0.001)} & 0.007 \Small{(0.001)} & 0.006 \Small{(0.001)} & \textbf{0.004} \Small{(0.000)} & 0.006 \Small{(0.000)} & 0.006 \Small{(0.003)}\\
 & SSUF3 & \textit{0.004} & 0.012 \Small{(0.001)} & 0.019 \Small{(0.006)} & 0.011 \Small{(0.002)} & \textbf{0.005} \Small{(0.001)} & 0.007 \Small{(0.002)} & 0.063 \Small{(0.052)}\\

 \hline
 \hline
\multirow{6}{*}{\rotatebox[origin=c]{90}{\textbf{IGDX}} \rotatebox[origin=c]{90}{(decision space)}} 
 & Two-On-One & \textit{0.013} & \textbf{0.026} \Small{(0.001)} & 0.035 \Small{(0.003)} & 0.037 \Small{(0.002)} & 0.043 \Small{(0.012)} & 0.148 \Small{(0.118)} & 0.281 \Small{(0.164)}\\
 & SYM-PART1 & \textit{0.051} & \textbf{0.073} \Small{(0.007)} & \textbf{0.069} \Small{(0.003)} & 0.148 \Small{(0.024)} & 9.427 \Small{(1.520)} & 7.929 \Small{(2.343)} & 9.155 \Small{(2.748)}\\
 & SYM-PART2 & \textit{0.052} & \textbf{0.070} \Small{(0.006)} & 0.078 \Small{(0.003)} & 0.161 \Small{(0.026)} & 9.410 \Small{(1.082)} & 5.371 \Small{(1.964)} & 9.483 \Small{(2.191)}\\
 & SYM-PART3 & \textit{0.042} & \textbf{0.053} \Small{(0.004)} & 0.148 \Small{(0.209)} & 0.491 \Small{(0.369)} & 8.335 \Small{(3.050)} & 5.841 \Small{(1.892)} & 7.397 \Small{(1.965)}\\
 & SSUF1 & \textit{0.055} & \textbf{0.057} \Small{(0.001)} & 0.076 \Small{(0.008)} & 0.086 \Small{(0.006)} & 0.142 \Small{(0.040)} & 0.132 \Small{(0.022)} & 0.244 \Small{(0.065)}\\
 & SSUF3 & \textit{0.008} & \textbf{0.016} \Small{(0.003)} & 0.030 \Small{(0.009)} & 0.020 \Small{(0.006)} & 0.162 \Small{(0.077)} & 0.071 \Small{(0.048)} & 0.308 \Small{(0.109)}\\
 
\hline
\hline
\end{tabular}
\end{center}
\end{table*}

\subsection{Benchmark comparison with budget}
Next, we benchmark {\alg} in the same experiment setup and budget as in \cite{tanabe18}, which allows $30\,000$ function evaluations and approximation set size of at most $N_\cA = 100$ solutions. {From the proposed problems, the Omni-Test \cite{deb05} was omitted, as it has $3^5 = 243$ global Pareto sets, which is more than the number of solutions allowed in the current experiment setup. Therefore, higher IGDX scores would be achieved when each solution is a non-Pareto optimal solution that is in between Pareto sets. This is not in line with the purpose of multi-modal optimization.} The IGD and IGDX are used as performance measures. The MR is omitted because the limited approximation set size $N_\cA$ does not allow all modes to be obtained simultaneously with a sufficient number of solutions to achieve the desired accuracy of $\varepsilon = 0.05$. Therefore, higher MR scores would be obtained by algorithms that obtain only a subset of the modes, which is again not the purpose of this experiment. {\alg} is equipped with MAM, based on its better IGDX performance on the MinDist problems shown in Figure~\ref{fig:convergence_lops}. The MMOEAs MO\_Ring\_PSO\_SCD  \cite{yue17} and MOEA/D-AD \cite{tanabe18}, and the MOEAs {\mam}, NSGA-II \cite{deb02}, and MOEA/D \cite{tanabe18} are used for comparison. We set the elitist archive size to $N_\bbE = 1000$. The approximation set is post-processed using greedy scattered subset selection in \textit{decision} space. The multi-start scheme was not used for {\alg} in this experiment as it is used to prevent the need to set the population size, but generally lowers performance in a limited budget setting as considered. Similar to the compared algorithms, {\alg} was ran with the standard population size recommended in literature. The population size $N$ and number of clusters is fixed according to the recommendations in \cite{rodrigues14} to $k = 20$ clusters and population size $N = \frac12 k \lfloor 17 +  3n^{\frac32}\rfloor = 250$, where $n$ is the number of decision variables. For {\mam}, the same setup as {\alg} is used, but the final elitist archive is post-processed with greedy scattered subset selection in \textit{objective} space. Obtained scores are tested for significance with the Wilcoxon rank-sum test with Bonferroni correction for the 60 tests performed, resulting in a significance level of $\alpha = \frac{0.01}{60} = 0.00017$.


\subsubsection{Results.}
Table~\ref{tab:igd} shows that {\mam} comes very close to the best achievable IGD for all problems, outperforming all other algorithms. This suggests that {\mam} is a sensible choice to use as a starting point for {\alg}. {\alg} outperforms all algorithms in terms of the IGDX on all problems but SYM-PART1. In all cases, {\alg} obtained all nine Pareto sets of the SYM-PART problems, which becomes increasingly harder for respectively SYM-PART1, 2, and 3. The reference Pareto set of the Two-On-One problem is an approximation that is accurate up to 0.0045 \cite{preuss06}, and multiple algorithms have IGD and IGDX scores close to this accuracy, which may obfuscate true performance.

\section{Discussion}
\label{sec:discussion}
We showed that {\hvc} succesfully distinguishes multi-objective niches, which are then explored separately in {\alg}. By considering Pareto domination per niche, local Pareto sets can be maintained. However, for problems with a large number of niches, (e.g. Omni-Test \cite{deb05}, SSUF1,  SSUF3), this approach results in a large number of clusters, and a very large population is required to explore all niches simultaneously. In that case, controlling the maximum number of clusters, or a serial search, as in \cite{maree18}, might be preferred.
 
Visualizing {\hvc} on problems with two decision variables provided useful insight in the behavior of {\hvc}. Performance of {\alg} on the MinDist problem with up to 20 variables shows that this technique extends to higher-dimensional problems. However, its performance and scalability has yet to be shown for more complex problems and even higher-dimensional decision spaces.

The current problem formulation of multi-modal optimization, where one is interested in only locating global Pareto sets, poses a fundamental difficulty. In practice, especially when a problem contains noise, the global Pareto front is never exactly attained, which makes it difficult or impossible to distinguish between global and high-quality local Pareto sets. 

It would furthermore be interesting to see how {\hvc} can be applied to improve performance when the aim is objective-space diversity in a multi-modal fitness landscape, rather than decision space diversity. 

By aiming {\alg} for decision-space diversity, objective-space diversity deteriorated for most problems, compared to {\mam}. This deterioration was found to relate mainly to the size limit of the elitist archive. If a balance between objective- and decision space diversity is desired, different size-control mechanisms could be applied to the elitist archive.

MOEA/D-AD \cite{tanabe18} is equipped with a population size growing scheme by which the population size is adapted to the problem at hand. In this work, a population-sizing scheme was used, but adapting this scheme or the population size itself on the number of detected niches could further improve performance. Both MOEA/D-AD and {\alg} use a post-processing step to construct a limited-size approximation set while maintaining a larger elitist archive during optimization, which seems to be beneficial.

\section{Conclusion}
\label{sec:conclusion}
In this work, we introduced {\hvc} for clustering multi-objective optimization problems into niches. We combined {\hvc} with {\mam} into the multi-modal multi-objective evolutionary algorithm {\alg}. We empirically show that {\alg} outperforms {\mam} and other multi-objective optimization algorithms in multi-modal optimization on a set of multi-modal benchmark functions. Furthermore, and perhaps most importantly, we show that {\alg} is capable of obtaining, maintaining, and exploiting multiple local Pareto sets simultaneously over time.

\begin{acks} 
This work is part of the research programme IPPSI-TA with project number 628.006.003, which is financed by the Netherlands Organisation for Scientific Research (NWO) and Elekta. We acknowledge financial support of the Nijbakker-Morra Foundation for a high-performance computing system. We thank N.H. Luong and P.A. Bouter for their support in the implementation of {\mam} in C++, and R. Tanabe for sharing the raw data of \cite{tanabe18}. 
\end{acks}
\bibliographystyle{ACM-Reference-Format}
\bibliography{Maree_2018c} 

\end{document}